\documentclass[11pt]{article}
\usepackage{paclic34}
\usepackage{times}
\usepackage{latexsym}
\usepackage{amsmath}
\usepackage{multirow}
\usepackage{url}
\usepackage{latexsym}
\usepackage{times}
\usepackage{url}
\usepackage{latexsym}
\usepackage{multirow}
\usepackage{amsmath}
\usepackage{adjustbox}
\usepackage{array}
\usepackage{subfig}
\usepackage{graphicx}
\usepackage{enumitem}
\usepackage{footnote}
\usepackage{booktabs}
\usepackage{xspace}
\makesavenoteenv{tabular}
\makesavenoteenv{table}

\newcolumntype{L}[1]{>{\raggedright\let\newline\\\arraybackslash\hspace{0pt}}m{#1}}
\newcolumntype{C}[1]{>{\centering\let\newline\\\arraybackslash\hspace{0pt}}m{#1}}
\newcolumntype{R}[1]{>{\raggedleft\let\newline\\\arraybackslash\hspace{0pt}}m{#1}}

\newcommand{\secref}[2][]{Section#1~\ref{sec:#2}}

\title{Towards Computational Linguistics in Minangkabau Language: \\Studies on Sentiment Analysis and Machine Translation}

\author{Fajri Koto \\
	School of Computing and Information System \\
	The University of Melbourne \\
	{\tt \small ffajri@student.unimelb.edu.au} \\\And
	Ikhwan Koto \\
	Faculty of IT\\
	Andalas University \\
	{\tt \small ikhwan220397@gmail.com} \\}

\date{}

\begin{document}
\maketitle
\begin{abstract}
  Although some linguists \cite{rusmali2005,crouch2009voice} have fairly attempted to define the morphology and syntax of Minangkabau, information processing in this language is still absent due to the scarcity of the annotated resource. In this work, we release two Minangkabau corpora: sentiment analysis and machine translation that are harvested and constructed from Twitter and Wikipedia.\footnote{Our data can be accessed at \url{https://github.com/fajri91/minangNLP}} We conduct the first computational linguistics in Minangkabau language employing classic machine learning and sequence-to-sequence models such as LSTM and Transformer. Our first experiments show that the classification performance over Minangkabau text significantly drops when tested with the model trained in Indonesian. Whereas, in the machine translation experiment, a simple word-to-word translation using a bilingual dictionary outperforms LSTM and Transformer model in terms of BLEU score.
\end{abstract}

\section{Introduction}
\label{sec:intro}


Minangkabau (Baso Minang) is an Austronesian language with roughly 7m speakers in the world \cite{gordon2005}. The language is spread under the umbrella of Minangkabau tribe -- a matrilineal culture in the province of West Sumatra, Indonesia. The first-language speakers of Minangkabau are scattered across Indonesian archipelago and Negeri Sembilan, Malaysia due to  ``\textit{merantau}" (migration) culture of Minangkabau tribe \cite{drakard1999a}.


Despite there being over 7m first-language speakers of Minangkabau,\footnote{In Indonesia, Minangkabau language is the fifth most spoken indigenous language after Javanese (75m), Sundanese (27m), Malay (20m), and Madurese (14m) \cite{riza2008resources}.} this language is rarely used in the formal sectors such as government and education. This is because the notion to use Bahasa Indonesia as the unity language since the Independent day of Indonesia in 1945 has been a double-edged sword. Today, Bahasa Indonesia successfully connects all ethnicities across provinces in Indonesia \cite{cohn2014local}, yet threatens the existent of some indigenous languages as the native speakers have been gradually decreasing \cite{novitasari2020cross}. \newcite{cohn2014local} predicted that Indonesia may shift into a monolingual society in the future.

In this paper, we initiate the preservation and the first information processing of Minangkabau language by constructing a Minangkabau--Indonesian parallel corpus, sourced from Twitter and Wikipedia. Unlike other indigenous languages such as Javanese and Sundanese that have been discussed in machine speech chain \cite{novitasari2020cross,wibawa2018building}, part-of-speech \cite{pratama2020part}, and translation system \cite{suryani2016enriching}, information processing in Minangkabau language is less studied. To the best of our knowledge, this is the first research on NLP in Minangkabau language, which we conduct in two different representative NLP tasks: sentiment analysis (classification) and machine translation (generation).

There are two underlying reasons why we limit our work in Minangkabau--Indonesian language pair. First, Minangkabau and Indonesian language are generally intelligible with some overlaps of lexicons and syntax. The Indonesian language has been extensively studied and is arguably a convenient proxy to learn the Minangkabau language. Second, authors of this work are the first-language speakers of Minangkabau and Indonesian language. This arguably eases and solidifies the research validation in both tasks.

To summarize, our contributions are: (1) we create a bilingual dictionary from Minangkabau Wikipedia by manually translating top 20,000 words into Indonesian; 2) we release Minangkabau corpus for sentiment analysis by manually translating 5,000 sentences of Indonesian sentiment analysis corpora; 3) We develop benchmark models with classic machine learning and pre-trained language model for Minangkabau sentiment analysis; 4) We automatically create a high-quality machine translation corpus consisting 16K Minangkabau--Indonesian parallel sentences; and 5) We showcase the first Minangkabau--Indonesian translation system through LSTM and Transformer model.


\section{Minangkabau--Indonesian Bilingual Dictionary}
\label{sec:dic}

\begin{table*}[t!]
	\begin{center}
		\begin{adjustbox}{max width=0.7\linewidth}
			\begin{tabular}{lll}
				\toprule
				\textbf{Indonesian} & \textbf{English} & \textbf {Minangkabau} \\
				\midrule
				\multicolumn{3}{c}{Synonyms} \\
				\midrule
				\textit{ibunya} & her mother & \textit{ibunyo, mandehnyo, amaknyo} \\
				\textit{memperlihatkan} & to show & \textit{mampacaliak, mampaliekan} \\
				\textit{kelapa} & coconut & \textit{karambia, kalapo} 
				\\
				\midrule
				\multicolumn{3}{c}{Dialectal variations} \\
				\midrule
				\textit{berupa} & such as & \textit{barupo, berupo, berupa, barupa} \\
				\textit{bersifat} & is, act, to have the quality	 & \textit{basipaik, basifaik, basifek, basifat, basipek} \\
				\textit{Belanda} & Netherlands & \textit{Balando, Belanda, Bulando, Belando} 
				\\
				\bottomrule
			\end{tabular}
		\end{adjustbox}
	\end{center}
	\caption{\label{tab:kamus} Example of synonyms and dialectal variations in the Minangkabau--Indonesian dictionary}
\end{table*}

In the province of West Sumatra, Minangkabau language is mostly used in spoken communication, while almost all reading materials such as local newspaper and books are written in Indonesian. Interestingly, Minangkabau language is frequently used in social media such as Twitter, Facebook and WhatsApp, although the writing can be varied and depends on the speaker dialect. \newcite{rusmali2005} define 6 Minangkabau dialects based on cities/regencies in the West Sumatra province. This includes Agam, Lima Puluh Kota, Pariaman, Tanah Datar, Pesisir Selatan, and Solok. The variation among these dialects is mostly phonetic and rarely syntactic. 

\newcite{crouch2009voice} classifies Minangkabau language into two types: 1) Standard Minangkabau and 2) Colloquial Minangkabau. The first type is the standard form for intergroup communication in the province of West Sumatra, while the second is the dialectal variation and used in informal and familiar contexts. \newcite{moussay1998} and \newcite{crouch2009voice} argue that Padang dialect is the standard form of Minangkabau. However, as the first-language speaker, we contend that these statements are inaccurate because of two reasons. 
First, many locals do not aware of Padang dialect. We randomly survey 28 local people and only half of them know the existence of Padang dialect. Second, in 2015 there has been an attempt to standardize Minangkabau language by local linguists, and Agam-Tanah Datar is proposed as the standard form due to its largest population.\footnote{\url{https://id.wikimedia.org/wiki/Sarasehan_Bahasa_Minangkabau}}

Our first attempt in this work is to create a publicly available Minangkabau--Indonesian dictionary by utilising Wikipedia. Minangkabau Wikipedia\footnote{Downloaded in June 2020} has 224,180 articles (rank 43rd) and contains 121,923 unique words, written in different dialects. We select top-20,000 words and manually translate it into Indonesian. We found that this collection contains many noises (e.g. scientific terms, such as \textit{onthophagus}, \textit{molophilus}) that are not Minangkabau nor Indonesian language. After manually translating the Minangkabau words, we use \textit{Kamus Besar Bahasa Indonesia} (KBBI)\footnote{\url{https://github.com/geovedi/indonesian-wordlist}} -- the official dictionary of Indonesian language to discard the word pairs with the unregistered Indonesian translation.
We finally obtain 11,905-size Minangkabau--Indonesian bilingual dictionary, that is 25 times larger than word collection in Glosbe (476 words).\footnote{\url{https://glosbe.com/min/id}}

We found that 6,541 (54.9\%) Minangkabau words in the bilingual dictionary are the same with the translation. As both Minangkabau and Indonesian languages are Austronesian (\textit{Malayic}) language, the high ratio of lexicon overlap is very likely. Further, we observe that 1,762 Indonesian words have some Minangkabau translations. These are primarily synonyms and dialectal variation that we show in Table~\ref{tab:kamus}. Next, in this study, we use this dictionary in sentiment analysis and machine translation.




\section{Sentiment Analysis}

Sentiment analysis has been extensively studied in English and Indonesia in different domains such as movie review \cite{yessenov2009sentiment,nurdiansyah2018sentiment}, Twitter \cite{agarwal2011sentiment,koto2015a}, and presidential election \cite{wang2012a,ibrahim2015buzzer}. It covers a wide range of approaches, from classic machine learning such as naive Bayes \cite{nurdiansyah2018sentiment}, SVM \cite{koto2015the} to pre-trained language models \cite{sun2019utilizing,xu2019bert}. The task is not only limited to binary classification of positive and negative polarity, but also multi classification \cite{liu2015a}, subjectivity classification \cite{liu2010sentiment}, and aspect-based sentiment \cite{ma2017interactive}.

In this work, we conduct a binary sentiment classification on positive and negative sentences by first manually translating Indonesian sentiment analysis corpus to Minangkabau language (Agam-Tanah Datar dialect). To provide a comprehensive preliminary study, we experimented with a wide range of techniques, starting from classic machine learning algorithms, recurrent models, to the state of the art technique, Transformer \cite{vaswani2017attention}.

\subsection{Dataset}


The data we use in this work is sourced from 1) \newcite{koto2017inset}; and 2) an aspect-based sentiment corpus.\footnote{\url{https://github.com/annisanurulazhar/absa-playground/}} \newcite{koto2017inset} dataset is originally from Indonesian tweets and has been labelled with positive and negative class. The second dataset is a hotel review collection where each review can encompass multi-polarity on different aspects. We determine the sentiment class based on the majority count of the sentiment label, and simply discard it if there is a tie between positive and negative. In total, we obtain 5,000 Indonesian texts from these two sources. We then ask two native speakers of Minangkabau and Indonesian language to manually translate all texts in the corpus. Finally, we create a parallel sentiment analysis corpus with 1,481 positive and 3,519 negative labels.


\subsection{Experimental Setup}

We conducted two types of the zero-shot experiment by using Indonesian train and development sets. In the first experiment, the model is tested against Minangkabau data, while in the second experiment we test the same model against the Indonesian translation, obtained by word-to-word translation using the bilingual dictionary (\secref{dic}). There are two underlying reasons to perform the zero-shot learning: 1) Minangkabau is intelligible with Indonesian language and most available corpus in the West Sumatra is Indonesian; 2) Minangkabau language is often mixed in Indonesian data collection especially in social media (e.g. Twitter, if the collection is based on geographical filter). Through zero-shot learning, we aim to measure the performance drop of Indonesian model when tested against the indigenous language like Minangkabau. 

Our experiments in this section are based on 5-folds cross-validation. We conduct stratified sampling with ratio 70/10/20 for train, development, and test respectively, and utilize five different algorithms as shown in Table~\ref{tab:res_sent}. For naive Bayes, SVM and logistic regression, we use byte-pair encoding (unigram and bigram) during the training and tune the model based on the development set. Due to data imbalance, we report the averaged F-1 score of five test sets.

For Bi-LSTM (200-d hidden size) we use two variants of 300-d word embedding: 1) random initialization; and 2) \texttt{fastText} pre-trained Indonesian embeddings \cite{bojanowski2016enriching}. First, we lowercase all characters and truncate them by 150 maximum words. We use batch size 100, and concatenate the last hidden states of Bi-LSTM for classification layer. For each fold, we train and tune the model for 100 steps with Adam optimizer and early stopping (patience = 20). 

Lastly, we incorporate the Transformer-based language model BERT \cite{devlin2019bert} in our experiment. Multilingual BERT (\textsc{mBERT}) is a masked language model trained by concatenating 104 languages in Wikipedia, including Minangkabau. \textsc{mBERT} has been shown to be effective for zero-shot cross-lingual tasks including classification \cite{wu2019beto}. In this work, we show the first utility of \textsc{mBERT} for classifying text in the indigenous language, such as Minangkabau. In fine-tuning, we truncate all data by 200 maximum tokens, and use batch size 30 and maximum epoch 20 (2,500 steps). The initial learning rate is 5e-5 with warm-up of 10\% of the total steps. We evaluate F-1 score of the development set for every epoch, and terminate the training if the performance does not increase within 5 epochs. Similar to Bi-LSTM models, we use Adam optimizer for gradient descent steps.

\begin{table}[t!]
	\begin{center}
		\begin{adjustbox}{max width=1\linewidth}
			\begin{tabular}{lccc}
				\toprule
				\multirow{2}{*}{\textbf{Method}} & \multicolumn{2}{c}{\textbf{Train\textsubscript{ ID}}} & \textbf{Train\textsubscript{ MIN}}  \\
				& \textbf{Test\textsubscript{ MIN}} & \textbf{Test\textsubscript{ ID'}} &  \textbf{Test\textsubscript{ MIN}}\\
				\midrule
				Naive Bayes	& 68.49 & 68.86 & 73.03 \\
				SVM	& 59.75 &	68.35 & 74.05 \\
				Logistic Regression & 57.95 & 66.90 & 72.35 \\
				Bi-LSTM & 58.75 & 65.62 & 72.37 \\
				Bi-LSTM + \texttt{fastText} & 62.06 & 71.51 & 70.47 \\
				\textsc{mBERT} & 62.71 & 67.60 & \textbf{75.91} \\
				\bottomrule
			\end{tabular}
		\end{adjustbox}
	\end{center}
	\caption{\label{tab:res_sent} Results for Sentiment Analysis on Minangkabau test set. The numbers are the averaged F-1 of 5-folds cross validation sets. MIN = Minangkabau, ID = Indonesian, ID' = Indonesian translation through bilingual dictionary.}
\end{table}

\subsection{Result}

In Table~\ref{tab:res_sent}, we show three different experimental results. The first column is the zero-shot setting where the model is trained and tuned using Indonesian text and tested against Minangkabau data. Surprisingly, naive Bayes outperforms other models including \textsc{mBERT} with a wide margin. Naive Bayes achieves 68.49 F1-score, +6 points over the pre-trained language model and Bi-LSTM + \texttt{fastText}. This might indicate that naive Bayes can effectively exploit the vocabulary overlap between Minangkabau and Indonesian language. 

In the second experiment, we hypothesize that a simple word-to-word translation using a bilingual dictionary can improve zero-shot learning. Similar to the first experiment, we train the model with Indonesian text, but we test the model against the Indonesian translation. As expected, the F-1 scores improve dramatically for all methods except naive Bayes with +0.37 gains. SVM, logistic regression and Bi-LSTMs are improved by 6--9 points while \textsc{mBERT} gains by +5 points by predicting the Indonesian translation.

In the third experiment, we again show a dramatic improvement when the model is fully trained in the Minangkabau language. Compared to the second experiment, all models are improved by 4--8 points with Bi-LSTM + \texttt{fastText} in exception. This is because the model uses \texttt{fastText} pre-trained Indonesian embeddings, and its best utility is when the model is trained and tested in the Indonesian language (second experiment). The best model is achieved by \textsc{mBERT} with 75.91 F1-score, outperforming other models with a comfortable margin. 

Based on these experiments, we can conclude the necessity of specific indigenous language resource for text classification in Indonesia. These languages are mixed in Indonesian social media, and testing the Indonesian model directly on this Indonesian-type language can drop the sentiment classification performance by 11.41 on average.

\subsection{Error Analysis}

In this section, we manually analyze the false positive (FN) and false negative (FP) of \textsc{mBERT} model. We examine all misclassified instances in the test set by considering three factors:
\begin{itemize}[noitemsep]
	\item \textit{Bias towards a certain topic}.
	In Indonesia, we argue that public sentiment towards government, politics and some celebrities are often negative. This could lead to bias in the training and result in a wrong prediction in the test set. We count the number of texts in FP and FN set that contain these two topics: politics and celebrity.
	\item \textit{Single polarity but containing words in opposite polarity}. The model might fail to correctly predict a sentiment label when contains words with the opposite polarity.
	\item \textit{Mixed polarity}. A text can consist of both positive and negative polarity with one of them is more dominant. 
\end{itemize}

\begin{table}[t]
	\begin{center}
		\begin{adjustbox}{max width=0.9\linewidth}
			\begin{tabular}{rrr}
				\toprule
				\bf Category & \bf Value &  \\
				\midrule
				\#FN & 83 & \\
				Bias towards certain topic (\%) & 34.84 & \\
				Single polarity with negative words (\%) & 20.48 & \\
				Mixed polarity (\%) & 12.05 & \\
				\midrule
				\#FP & 56 & \\
				Bias towards certain topic (\%) & 26.79 & \\
				Single polarity with positive words (\%) & 26.79 & \\
				Mixed polarity (\%) & 28.57 & \\
				\bottomrule
			\end{tabular}
		\end{adjustbox}
	\end{center}
	\caption{\label{tab:err} Error analysis for False Negative (FN) and False Positive (FP) set.	}
\end{table}

\begin{figure*}[t!]
	\centering
	\includegraphics[width=5.5in]{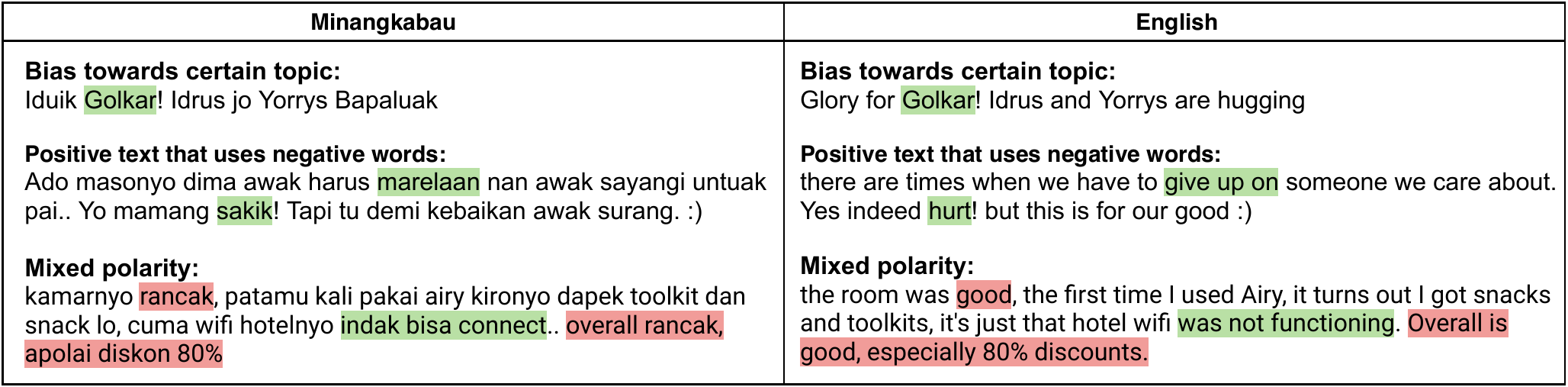}
	\caption{\label{fig:example} Example of False Negative.}
\end{figure*}

In Table~\ref{tab:err} we found that there are 83 FN (28\% of positive data) and  56 FP (8\% of negative data) instances. We further observe that 34.84\% of FN instances contain politics or celebrity topic, while there is only 26.79\% of FP instances with these criteria. In Figure~\ref{fig:example}, we show an FN example for the first factor: ``\textit{Iduik Golkar! Idrus jo Yorrys Bapaluak}'' where ``\textit{Golkar}'' is one of the political parties in Indonesia. 

Secondly, 20.48\% of FN instances contain negative words. As shown in Figure~\ref{fig:example} the example uses words ``give up'' and ``hurt'' to convey positive advice. We notice that the second factor is more frequent in FP instances with 26.79\% proportion. Lastly, we find that 28.57\% of FP instances have mixed polarity, 2 times larger than FN. We observe that most samples with mixed polarity are sourced from the hotel review. It highlights that mixed polarity is arguably a harder task, and requires special attention to aspects in text fragments.

\section{Machine Translation}

\begin{figure*}[t!]
	\centering
	\includegraphics[width=5.7in]{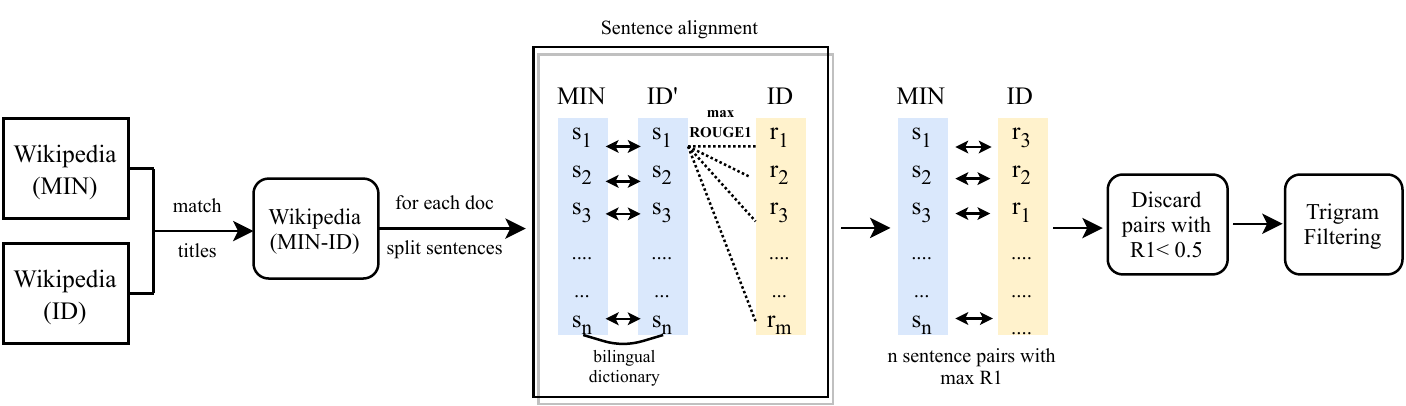}
	\caption{\label{fig:min-id} Flow chart of MIN-ID parallel corpus construction.}
\end{figure*}

Machine translation has been long run research, started by Rule-based Machine Translation (RBMT) \cite{carbonell1978,nagao1984a}, Statistical Machine Translation (SMT) \cite{brown1990a}, to Neural Machine Translation (NMT) \cite{bahdanau2015neural}. NMT with its continuous vector representation has been a breakthrough in machine translation, minimizing the complexity of SMT yet boosts the BLEU score \cite{papineni2002bleu} into a new level. Recently, \newcite{hassan2018achieving} announce that their Chinese--English NMT system has achieved a comparable result with human performance. 

Although there are 2,300 languages across Asia, only some Asian languages such as Chinese and Japanese have been extensively studied for machine translation. We argue there are two root causes: a lack of parallel corpus, and a lack of resource standardization. Apart from the Chinese language, there have been some attempts to create a parallel corpus across Asian languages. \newcite{nomoto2019interpersonal} construct 1,3k parallel sentences for Japanese, Burmese, Malay, Indonesian, Thai, and Vietnamese, while \newcite{kunchukuttan2017iit} release a large-scale Hindi-English corpus. Unlike these national languages, machine translation on indigenous languages is still very rare due to data unavailability. In Indonesia, Sundanese \cite{suryani2015experiment} and Javanese \cite{wibawa2013indonesian} have been explored through statistical machine translation. In this work, our focus is Minangkabau--Indonesian language pair, and we first construct the translation corpus from Wikipedia.

\subsection{Dataset}

Constructing parallel corpus through sentence alignment from bilingual sources such as news \cite{zhao2002adaptive,rauf2011parallel}, patent \cite{utiyama2003reliable,lu2010building}, and Wikipedia \cite{yasuda2008method,smith2010extracting,chu2014constructing} have been done in various language pairs. For Indonesian indigenous language, \newcite{trisedya2014creating} has attempted to create parallel Javanese--Indonesian corpus by utilizing inter-Wiki links and aligning sentences via \newcite{gale1993a} approach. 

In this work, we create Minangkabau--Indonesian (MIN-ID) parallel corpus by using Wikipedia\footnote{Downloaded in June 2020} (Figure~\ref{fig:min-id}). We obtain 224,180  Minangkabau and 510,258 Indonesian articles, and align documents through title matching, resulting in 111,430 MIN-ID document pairs. After that, we do sentence segmentation based on simple punctuation heuristics and obtain 4,323,315 Minangkabau sentences. We then use the bilingual dictionary (Section 2) to translate Minangkabau article (MIN) into Indonesian language (ID'). Sentence alignment is conducted using ROUGE-1 (F1) score (unigram overlap) \cite{lin2004rouge} between ID' and ID, and we pair each MIN sentence with an ID sentence based on the highest ROUGE-1. We then discard sentence pairs with a score of less than 0.5 to result in 345,146 MIN-ID parallel sentences.

We observe that the sentence pattern in the collection is highly repetitive (e.g. 100k sentences are about biological term definition). Therefore, we conduct final filtering based on top-1000 trigram by iteratively discarding sentences until the frequency of each trigram equals to 100. Finally, we obtain 16,371 MIN-ID parallel sentences and conducted manual evaluation by asking two native Minangkabau speakers to assess the adequacy and fluency \cite{koehn2006manual}. The human judgement is based on scale 1--5 (1 means poor quality and 5 otherwise) and conducted against 100 random samples. We average the weights of two annotators before computing the overall score, and we achieve 4.98 and 4.87 for adequacy and fluency respectively.\footnote{The Pearson correlation of two annotators for adequacy and fluency are 0.9433 and 0.5812 respectively} This indicates that the resulting corpus is high-quality for machine translation training.

\begin{table}[t]
	\begin{center}
		\begin{adjustbox}{max width=1\linewidth}
			\begin{tabular}{rrrrr}
				\toprule
				\multirow{2}{*}{\bf Category} & \multicolumn{2}{c}{\bf Wiki} &   \multicolumn{2}{c}{\bf SentC} \\
				& \textbf{MIN} & \textbf{ID}& \textbf{MIN} & \textbf{ID} \\
				\midrule
				mean(\#word) & 19.6 & 19.6 & 22.3 & 22.2 \\
				std(\#word) & 11.6 & 11.5 &  12.8 & 12.7 \\
				mean(\#char) & 105.2 & 107.7 & 98.9 & 99.2 \\
				std(\#char) &  59.1 &  60.4 &  57.9 &  58.1 \\
				\#vocab & 32,420 & 27,318 & 13,940 & 13,698 \\
				Overlapping \#vocab & \multicolumn{2}{c}{21,563} & \multicolumn{2}{c}{9,508} \\
				\bottomrule
			\end{tabular}
		\end{adjustbox}
	\end{center}
	\caption{\label{tab:stat} Statistics of machine translation corpora.}
\end{table}

\subsection{Experimental Setup}

First, we split Wikipedia data with ratio 70/10/20, resulting in 11,571/1,600/3,200 data for train, development, and test respectively. In addition, we use parallel sentiment analysis corpus (Section 3) as the second test set (size 5,000) for evaluating texts from different domain. In Table~\ref{tab:stat}, we provide the overall statistics of both corpora: Wikipedia (Wiki) and Sentiment Corpus (SentC). We observe that Minangkabau (MIN) and Indonesian (ID) language generally have similar word and char lengths. The difference is in the vocabulary size where Minangkabau is 5k larger than Indonesian in Wiki corpus. As we discuss in Section 2, this difference is due to various Minangkabau dialects in Wikipedia.

We conducted two experiments: 1) Minangkabau to Indonesian (MIN $\rightarrow$ ID); and 2) Indonesian to Minangkabau (ID $\rightarrow$ MIN) with three models: 1) word-to-word translation (W2W) using bilingual dictionary (Section 2); 2) LSTMs; and 3) Transformer. We use Moses Tokeniser\footnote{\url{https://pypi.org/project/mosestokenizer/}} for tokenization, and sacreBLEU script \cite{post2018a} to evaluate BLEU score on the test sets. All source and target sentences are truncated by 75 maximum lengths.

Our encoder-decoder (LSTM and Transformer) models are based on Open-NMT implementation \cite{opennmt}. For LSTM models, we use two layers of 200-d Bi-LSTM encoder and 200-d LSTM decoder with a global attention mechanism. Source and target embeddings are 500-d and shared between encoder and decoder. For training, we set the learning rate of 0.001 with Adam optimizer, and warm-up of 10\% of the total steps. We train the model with batch size 64 for 50,000 steps and evaluate the development set for every 5,000 steps.

The Transformer encoder-decoder (each) has 6 hidden layers, 512 dimensionality, 8 attention heads, and 2,028 feed-forward dimensionalities. Similar to LSTM model, the word embeddings are shared between source and target text. We use cosine positional embedding and train the model with batch size 5,000 for 50,000 steps with Adam optimizer (warm-up = 5,000 and Noam decay scheme). We evaluate the development set for every 10,000 steps.

\subsection{Result}

\begin{table}[t]
	\begin{center}
		\begin{adjustbox}{max width=1\linewidth}
			\begin{tabular}{lrrrr}
				\toprule
				\multirow{2}{*}{\bf Method} & \multicolumn{2}{c}{\bf MIN $\rightarrow$ ID} &   \multicolumn{2}{c}{\bf ID $\rightarrow$ MIN} \\
				& \textbf{Wiki} & \textbf{SentC}& \textbf{Wiki} & \textbf{SentC} \\
				\midrule
				Raw (baseline) & 30.08 & 43.73 & 30.08 & 43.73 \\
				W2W & \bf 64.54 & \bf 60.99 & \bf 55.08 & \bf 55.22 \\
				LSTM & 63.77 & 22.82  & 48.50 & 15.52 \\
				Transformer & 56.25 & 10.23 & 43.50 & 8.86\\
				\bottomrule
			\end{tabular}
		\end{adjustbox}
	\end{center}
	\caption{\label{tab:bleu} BLEU score on the test set. SentC is parallel sentiment analysis corpus in Section 3.}
\end{table}

\begin{figure*}[t!]
	\centering
	\includegraphics[width=5.8in]{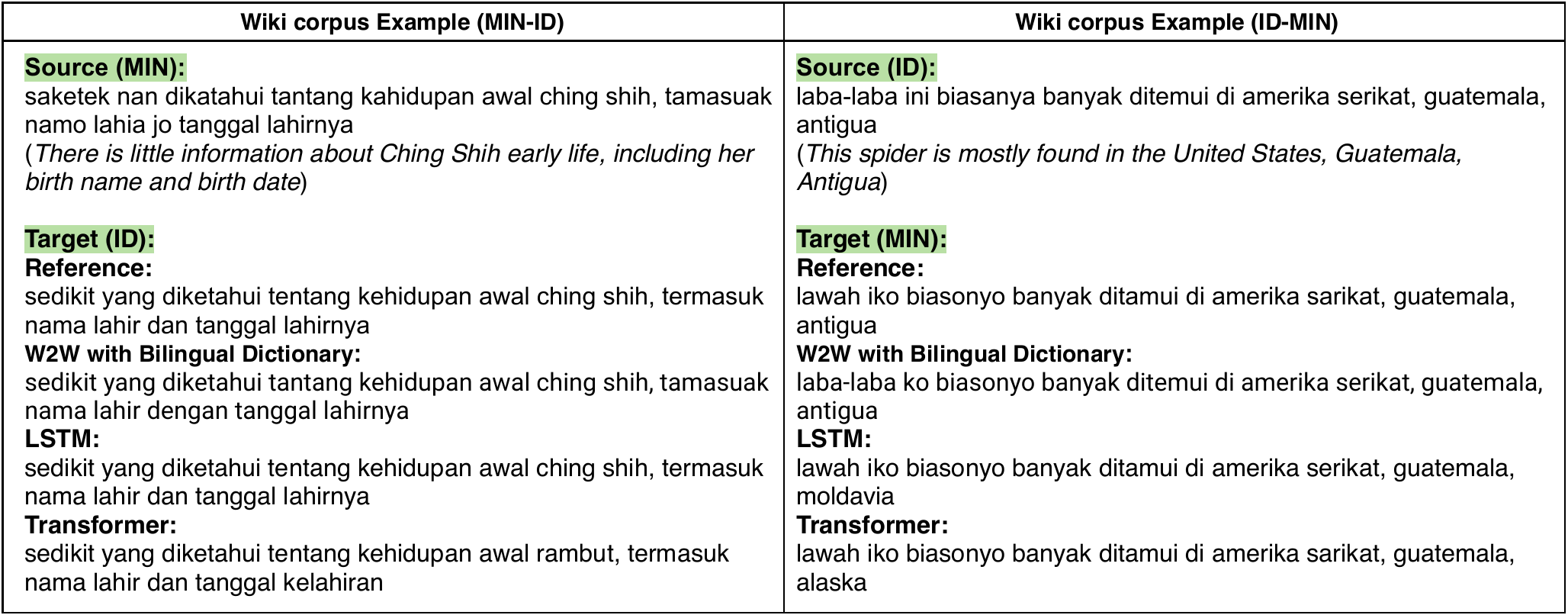}
	\caption{\label{fig:MIN} Examples of model translation in Wikipedia corpus.}
\end{figure*}

\begin{figure}[t!]
	\centering
	\includegraphics[width=3in]{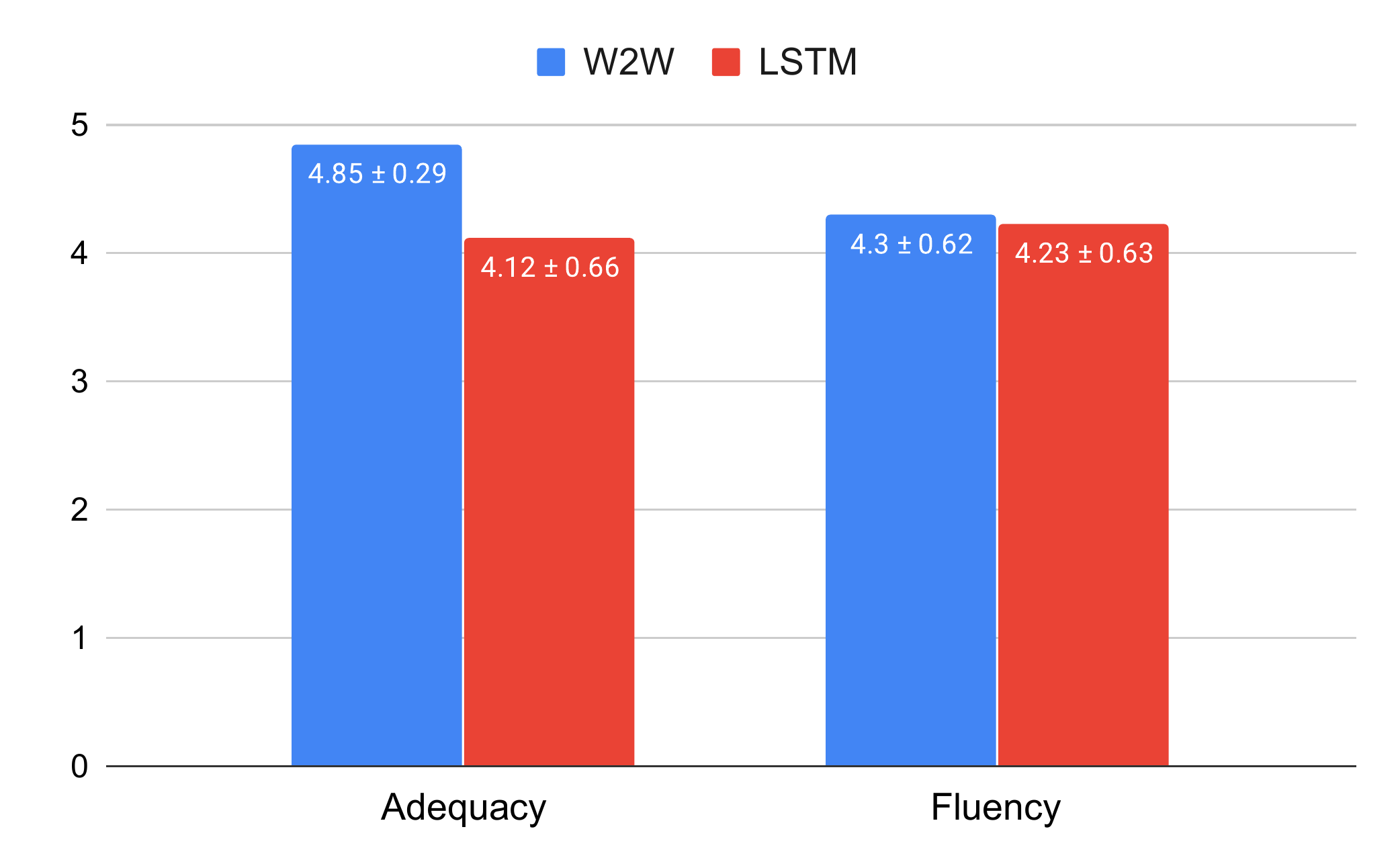}
	\caption{\label{fig:chart} Human judgment on MIN$\rightarrow$ID (Wiki)}
\end{figure}

In Table~\ref{tab:bleu}, we present the experiment results for machine translation. Because Indonesian and Minangkabau language is mutually intelligible, and  Table~\ref{tab:stat} shows that roughly 75\% words in two vocabularies overlap, we set the BLEU scores of raw source and target text as the baseline. We found for both MIN$\rightarrow$ID and ID$\rightarrow$MIN, the BLEU scores are relatively high, more than 30 points.

We observe that a simple word-to-word (W2W) translation using a bilingual MIN-ID dictionary achieves the best performance over LSTM and Transformer model in all cases. For MIN$\rightarrow$ID, the BLEU scores are 64.54 and 60.99 for Wiki and SentC respectively, improving the baseline roughly 20--30 points. The similar result is also found in ID$\rightarrow$MIN with disparity 12--25 points in the baseline. 

Both LSTM and Transformer models significantly improve the baseline for Wiki corpus, but poorly perform in translating SentC dataset. For the Wiki corpus, the LSTM model achieves a competitive score in MIN$\rightarrow$ID and ID$\rightarrow$MIN, improving the baseline for 33 and 18 points respectively. The Transformer also outperforms the baselines, but substantially lower than the LSTM. In out-of-domain test set (SentC), the performance of both models significantly drops, 20--30 points lower than the baselines. We further observe that this is primarily due to out of vocabulary issue, where around 65\% words in SentC are not in the vocabulary model.

\subsection{Analysis}

In Figure~\ref{fig:MIN}, we show translation examples in Wiki corpus. In MIN$\rightarrow$ID, the LSTM translation is slightly more eloquent than word-to-word (W2W) translation. Word ``\textit{tamasuak}'' (including) in W2W translation is Minangkabau language and not properly translated. This is because the word ``\textit{tamasuak}'' is not registered in the bilingual vocabulary. In this sample, Transformer model hallucinates as mentioning ``\textit{kehidupan awal rambut}'' (early life of hair), resulting in a poor fluency and adequacy. Next in ID$\rightarrow$MIN, W2W translation is better than LSTM and Transformer. W2W translation is relatively good, despite word ``\textit{ditemui}'' (found) that is not translated into Minangkabau. In this example, LSTM and Transformer hallucinate mentioning incorrect location such as ``\textit{moldavia}'', and ``\textit{alaska}''.

For further analysis, we conduct a manual evaluation on two best models: W2W and LSTM in MIN$\rightarrow$ID (Wiki) experiment. Like manual evaluation in Section 4.1, we ask two native Indonesian and Minangkabau speakers to examine the adequacy and fluency of 100 random samples with scale 1--5. Figure 1 shows that W2W translation significantly better than LSTM in terms of adequacy, but similar in terms of fluency. This is in line with our observation, that the LSTM model frequently generates incorrect keywords in a fluent and coherent translation. This is possibly due to the out-of-vocabulary (OOV) case in the test set, triggered by the small-size of our train set. A proper training scheme for the low-resource setting can be leveraged in future work, so it can reduce hallucination issue in the LSTM model.


\section{Conclusion}
In this work, we have shown the first NLP tasks in Minangkabau language. In sentiment analysis task, we found the necessity of indigenous language corpus for classifying Indonesian texts. Although Indonesian and Minangkabau languages are from \textit{Malayic} family, the Indonesian model can not optimally classify Minangkabau text. Next, in the machine translation experiment, although the word-to-word translation is superior to LSTM and Transformer, there is still a room of improvement for fluency. This can be addressed by training seq-to-seq model with a larger corpus.

\section*{Acknowledgments}

We are grateful to the anonymous reviewers for their
helpful feedback and suggestions. In this research, Fajri Koto is supported by the Australia Awards Scholarship (AAS), funded by Department of Foreign Affairs and Trade (DFAT) Australia.


\begin{thebibliography}{}

\bibitem[\protect\citename{Agarwal et al.}2011]{agarwal2011sentiment}
Apoorv {Agarwal}, Boyi {Xie}, Ilia {Vovsha}, Owen {Rambow}, and Rebecca {Passonneau}.
\newblock 2011.
\newblock Sentiment Analysis of Twitter Data.
\newblock {In \em Proceedings of the Workshop on Language in Social Media (LSM 2011)}.
\newblock pages 30--38.

\bibitem[\protect\citename{Bahdanau et al.}2015]{bahdanau2015neural}
Dzmitry {Bahdanau}, Kyunghyun {Cho}, and Yoshua {Bengio}.
\newblock 2015.
\newblock Neural Machine Translation by Jointly Learning to Align and Translate.
\newblock {In \em ICLR 2015 : International Conference on Learning Representations 2015}.

\bibitem[\protect\citename{Bojanowski et al.}2016]{bojanowski2016enriching}
Piotr Bojanowski, Edouard Grave, Armand  Joulin, Tomas Mikolov.
\newblock 2016.
\newblock Enriching Word Vectors with Subword Information.
\newblock {\em arXiv preprint arXiv:1607.04606}.

\bibitem[\protect\citename{Brown et al.}1990]{brown1990a}
\newblock Peter~F. {Brown}, John {Cocke}, Stephen~A.~Della {Pietra}, Vincent~J.~Della {Pietra}, Fredrick {Jelinek}, John~D. {Lafferty}, Robert~L. {Mercer}, and Paul~S. {Roossin}.
\newblock 1990.
\newblock A statistical approach to machine translation.
\newblock {\em A statistical approach to machine translation}, 16(2): 79--85.

\bibitem[\protect\citename{Carbonell et al.}1978]{carbonell1978}
\newblock Jaime~G {Carbonell}, Richard~E {Cullinford}, and Anatole~V {Gershman}.
\newblock 1978.
\newblock {\em Knowledge-based machine translation.}
\newblock {Technical report, Yale University, Department of Computer Science}, Connecticut, US.

\bibitem[\protect\citename{Chu et al.}2014]{chu2014constructing}
\newblock Chenhui {Chu}, Toshiaki {Nakazawa}, and Sadao {Kurohashi}.
\newblock 2014.
\newblock {Constructing a Chinese--Japanese Parallel Corpus from Wikipedia.}
\newblock {In \em Proceedings of the Ninth International Conference on Language Resources and Evaluation (LREC'14)}.
\newblock pages 642--647.

\bibitem[\protect\citename{Cohn et al.}2014]{cohn2014local}
\newblock Abigail~C Cohn, and Maya Ravindranath.
\newblock 2014.
\newblock {Local languages in Indonesia: Language maintenance or language shift.}
\newblock {\em Linguistik Indonesia}, 32(2): 131--148.

\bibitem[\protect\citename{Crouch}2009]{crouch2009voice}
\newblock Sophie~Elizabeth Crouch.
\newblock 2009.
\newblock {\em Voice and verb morphology in {M}inangkabau, a language of {W}est {S}umatra, Indonesia.}
\newblock {Master Thesis, The University of Western Australia.}

\bibitem[\protect\citename{Devlin et al.}2019]{devlin2019bert}
\newblock Jacob~Devlin, \xspace Ming-Wei~Chang, \xspace Kenton~Lee, \xspace\xspace and Kristina~Toutanova.
\newblock 2019.
\newblock {BERT: Pre-training of Deep Bidirectional Transformers for Language Understanding.}
\newblock {In \em NAACL-HLT 2019: Annual Conference of the North American Chapter of the Association for Computational Linguistics}.
\newblock pages 4171--4186.

\bibitem[\protect\citename{Drakard}1999]{drakard1999a}
\newblock Jane {Drakard}.
\newblock 1999.
\newblock {\em A Kingdom of Words: Language and Power in Sumatra.}

\bibitem[\protect\citename{Gale and Church}1993]{gale1993a}
\newblock William~A. \xspace {Gale}, \xspace and \xspace Kenneth \xspace W. \xspace {Church}.
\newblock 1993.
\newblock {A program for aligning sentences in bilingual corpora.}
\newblock {\em Computational Linguistic}, 19(1): 75--102.

\bibitem[\protect\citename{Gordon}2005]{gordon2005}
\newblock Raymond~G. Gordon.
\newblock 2005.
\newblock {\em Ethnologue: languages of the world, Fifteenth Edition.}
\newblock SIL International, Dallas, Texas.

\bibitem[\protect\citename{Hassan et al.}2018]{hassan2018achieving}
\newblock Hany {Hassan}, \xspace Anthony {Aue}, \xspace Chang {Chen}, \xspace Vishal {Chowdhary}, Jonathan {Clark}, Christian {Federmann}, Xuedong {Huang}, Marcin {Junczys-Dowmunt}, William {Lewis}, Mu {Li}, Shujie {Liu}, Tie-Yan {Liu}, Renqian {Luo}, Arul {Menezes}, Tao {Qin}, Frank {Seide}, Xu {Tan}, Fei {Tian}, Lijun {Wu}, Shuangzhi {Wu}, Yingce {Xia}, Dongdong {Zhang}, Zhirui {Zhang}, and Ming {Zhou}.
\newblock 2018.
\newblock {Achieving Human Parity on Automatic Chinese to English News Translation.}
\newblock {\em arXiv preprint arXiv:1803.05567}.

\bibitem[\protect\citename{Ibrahim et al.}2015]{ibrahim2015buzzer}
\newblock Mochamad {Ibrahim}, Omar {Abdillah}, Alfan F. {Wicaksono}, and Mirna {Adriani}.
\newblock 2015.
\newblock {Buzzer Detection and Sentiment Analysis for Predicting Presidential Election Results in a Twitter Nation.}
\newblock {In \em 2015 IEEE International Conference on Data Mining Workshop (ICDMW)}.
\newblock pages 1348--1353.

\bibitem[\protect\citename{Klein et al.}2017]{opennmt}
\newblock Guillaume Klein, Yoon Kim, Yuntian Deng, Jean Senellart, and Alexander~M. Rush.
\newblock 2017.
\newblock {Open{NMT}: Open-Source Toolkit for Neural Machine Translation.}
\newblock {In \em Proc. ACL}.

\bibitem[\protect\citename{Koehn and Monz}2006]{koehn2006manual}
\newblock Philipp {Koehn} and Christof {Monz}.
\newblock 2006.
\newblock {Manual and Automatic Evaluation of Machine Translation between European Languages.}
\newblock {In \em Proceedings on the Workshop on Statistical Machine Translation}.
\newblock pages 102--121.

\bibitem[\protect\citename{Koto and Adriani}2015]{koto2015a}
\newblock Fajri {Koto} and Mirna {Adriani}.
\newblock 2015.
\newblock {A Comparative Study on Twitter Sentiment Analysis: Which Features are Good?}
\newblock {In \em 20th International Conference on Applications of Natural Language to Information Systems, NLDB 2015}.
\newblock pages 453--457.

\bibitem[\protect\citename{Koto and Adriani}2015]{koto2015the}
\newblock Fajri {Koto} and Mirna {Adriani}.
\newblock 2015.
\newblock {The Use of POS Sequence for Analyzing Sentence Pattern in Twitter Sentiment Analysis}.
\newblock {In \em 2015 IEEE 29th International Conference on Advanced Information Networking and Applications Workshops}.
\newblock pages 547--551.

\bibitem[\protect\citename{Koto and Rahmaningtyas}2017]{koto2017inset}
\newblock Fajri {Koto} and Gemala Y. {Rahmaningtyas}.
\newblock 2017.
\newblock {InSet lexicon: Evaluation of a word list for Indonesian sentiment analysis in microblogs}.
\newblock {In \em 2017 International Conference on Asian Language Processing (IALP)}.
\newblock pages 391--394.

\bibitem[\protect\citename{Kunchukuttan et al.}2017]{kunchukuttan2017iit}
\newblock Anoop Kunchukuttan, Pratik Mehta, and Pushpak Bhattacharyya.
\newblock 2017.
\newblock {The IIT Bombay English-Hindi parallel corpus}.
\newblock {\em arXiv preprint arXiv:1710.02855}.

\bibitem[\protect\citename{Lin}2004]{lin2004rouge}
\newblock Chin-Yew {Lin}.
\newblock 2004.
\newblock {{ROUGE}: A Package for Automatic Evaluation of Summaries}.
\newblock {In \em Text Summarization Branches Out: Proceedings of the ACL-04 Workshop}.
\newblock pages 74--81.

\bibitem[\protect\citename{Liu and Chen}2015]{liu2015a}
\newblock Shuhua Monica {Liu} and Jiun-Hung {Chen}.
\newblock 2015.
\newblock {A multi-label classification based approach for sentiment classification}.
\newblock {\em Expert Systems With Applications}, 42(3): 1083--1093

\bibitem[\protect\citename{Liu}2010]{liu2010sentiment}
\newblock Bing {Liu}.
\newblock 2010.
\newblock {Sentiment Analysis and Subjectivity}.
\newblock {In \em Handbook of Natural Language Processing}.
\newblock pages 627--666.

\bibitem[\protect\citename{Lu et al.}2010]{lu2010building}
\newblock Bin {Lu}, Tao {Jiang}, Kapo {Chow}, and Benjamin~K. {Tsou}.
\newblock 2010.
\newblock {Building a Large English-Chinese Parallel Corpus from Comparable Patents and its Experimental Application to SMT}.

\bibitem[\protect\citename{Ma et al.}2017]{ma2017interactive}
\newblock Dehong {Ma}, Sujian {Li}, Xiaodong {Zhang}, and Houfeng {Wang}.
\newblock 2017.
\newblock {Interactive attention networks for aspect-level sentiment classification}.
\newblock {In \em IJCAI'17 Proceedings of the 26th International Joint Conference on Artificial Intelligence}.
\newblock pages 4068--4074.

\bibitem[\protect\citename{Moussay}1998]{moussay1998}
\newblock G\'{e}rard Moussay.
\newblock 1998.
\newblock {Tata Bahasa Minangkabau}.
\newblock {Kepustakaan Populer Gramedia, Jakarta}.

\bibitem[\protect\citename{Nagao}1984]{nagao1984a}
\newblock Makoto {Nagao}.
\newblock 1984.
\newblock {A framework of a mechanical translation between Japanese and English by analogy principle}.
\newblock {In \em Proc. of the international NATO symposium on Artificial and human intelligence}.
\newblock pages 173--180.

\bibitem[\protect\citename{Nomoto et al.}2019]{nomoto2019interpersonal}
\newblock Hiroki Nomoto, Kenji Okano, Sunisa Wittayapanyanon, and Junta Nomura.
\newblock 2019.
\newblock {Interpersonal meaning annotation for Asian language corpora: The case of TUFS Asian Language Parallel Corpus (TALPCo)}.
\newblock {In \em Proceedings of the Twenty-Fifth Annual Meeting of the Association for Natural Language Processing}.
\newblock pages 846--849.

\bibitem[\protect\citename{Novitasari et al.}2020]{novitasari2020cross}
\newblock Sashi {Novitasari}, Andros {Tjandra}, Sakriani {Sakti}, and Satoshi {Nakamura}.
\newblock 2020.
\newblock {Cross-Lingual Machine Speech Chain for Javanese, Sundanese, Balinese, and Bataks Speech Recognition and Synthesis}.
\newblock {In \em SLTU/CCURL@LREC}.
\newblock pages 131--138.

\bibitem[\protect\citename{Nurdiansyah et al.}2018]{nurdiansyah2018sentiment}
\newblock Yanuar {Nurdiansyah}, Saiful {Bukhori}, and Rahmad {Hidayat}.
\newblock 2018.
\newblock {Sentiment Analysis System for Movie Review in {B}ahasa {I}ndonesia using Naive Bayes Classifier Method}.
\newblock {In \em Journal of Physics: Conference Series}.
\newblock pages 12011.

\bibitem[\protect\citename{Papineni et al.}2002]{papineni2002bleu}
\newblock Kishore {Papineni}, Salim {Roukos}, Todd {Ward}, and Wei-Jing {Zhu}.
\newblock 2002.
\newblock {BLEU: a Method for Automatic Evaluation of Machine Translation}.
\newblock {In \em Proceedings of 40th Annual Meeting of the Association for Computational Linguistics}.
\newblock pages 311--318.

\bibitem[\protect\citename{Post}2018]{post2018a}
\newblock Matt {Post}.
\newblock 2018.
\newblock {A Call for Clarity in Reporting BLEU Scores}.
\newblock {In \em Proceedings of the Third Conference on Machine Translation: Research Papers}.
\newblock pages 186--191.

\bibitem[\protect\citename{Pratama et al.}2020]{pratama2020part}
\newblock Ryan~Armiditya Pratama, Arie~Ardiyanti Suryani, and Warih Maharani.
\newblock 2020.
\newblock {Part of Speech Tagging for Javanese Language with Hidden Markov Model}.
\newblock {In \em Journal of Computer Science and Informatics Engineering (J-Cosine)}, 4(1): 84--91.

\bibitem[\protect\citename{Rauf and Schwenk}2011]{rauf2011parallel}
\newblock Sadaf Abdul {Rauf} and Holger {Schwenk}.
\newblock 2011.
\newblock {Parallel sentence generation from comparable corpora for improved SMT}.
\newblock {In \em Machine Translation}, 25(4): 341--375.

\bibitem[\protect\citename{Riza}2008]{riza2008resources}
\newblock Hammam {Riza}.
\newblock 2008.
\newblock {Resources Report on Languages of Indonesia}.
\newblock {In \em ALR@IJCNLP}.
\newblock pages 93--94.

\bibitem[\protect\citename{Rusmali et al.}1985]{rusmali2005}
\newblock Marah Rusmali, Amir~Hakim Usman, Syahwin Nikelas, Nurzuir Husin, Busri Busri, Agusli Lana, M. Yamin, Isna Sulastri, and Irfani Basri.
\newblock 1985.
\newblock {Kamus {M}inangkabau -- {I}ndonesia}.
\newblock {Pusat Pembinaan dan Pengembangan Bahasa Departemen Pendidikan dan Kebudayaan, Jakarta}.

\bibitem[\protect\citename{Smith et al.}2010]{smith2010extracting}
\newblock Jason~R. {Smith}, Chris {Quirk}, and Kristina {Toutanova}.
\newblock 2010.
\newblock {Extracting Parallel Sentences from Comparable Corpora using Document Level Alignment}.
\newblock {In \em Human Language Technologies: The 2010 Annual Conference of the North American Chapter of the Association for Computational Linguistics}.
\newblock pages 403--411.

\bibitem[\protect\citename{Sun et al.}2019]{sun2019utilizing}
\newblock Chi~{Sun}, Luyao {Huang}, and Xipeng {Qiu}.
\newblock 2019.
\newblock {Utilizing {BERT} for Aspect-Based Sentiment Analysis via Constructing Auxiliary Sentence}.
\newblock {In \em NAACL-HLT}.
\newblock pages 380--385.

\bibitem[\protect\citename{Suryani et al.}2015]{suryani2015experiment}
\newblock Arie~Ardiyanti {Suryani}, Dwi~Hendratmo {Widyantoro}, Ayu {Purwarianti}, and Yayat {Sudaryat}.
\newblock 2015.
\newblock {Experiment on a phrase-based statistical machine translation using POS Tag information for Sundanese into Indonesian}.
\newblock {In \em 2015 International Conference on Information Technology Systems and Innovation (ICITSI)}.
\newblock pages 1--6.

\bibitem[\protect\citename{Suryani et al.}2016]{suryani2016enriching}
\newblock Arie \xspace Ardiyanti \xspace {Suryani}, \xspace Isye \xspace Arieshanti, \xspace Banu W. Yohanes, M. Subair, Sari D.  Budiwati, and Bagus S. Rintyarna.
\newblock 2016.
\newblock {Enriching English into Sundanese and Javanese translation list using pivot language}.
\newblock {In \em 2016 International Conference on Information \& Communication Technology and Systems (ICTS)}.
\newblock pages 167--171.

\bibitem[\protect\citename{Trisedya and Inastra}2014]{trisedya2014creating}
\newblock Bayu \xspace Distiawan \xspace {Trisedya} \xspace and Dyah \xspace {Inastra}.
\newblock 2014.
\newblock {Creating Indonesian-Javanese parallel corpora using wikipedia articles}.
\newblock {In \em 2014 International Conference on Advanced Computer Science and Information System}.
\newblock pages 239--245.

\bibitem[\protect\citename{Utiyama and Isahara}2003]{utiyama2003reliable}
\newblock Masao {Utiyama} and Hitoshi {Isahara}.
\newblock 2003.
\newblock {Reliable Measures for Aligning Japanese-English News Articles and Sentences}.
\newblock {In \em Proceedings of the 41st Annual Meeting of the Association for Computational Linguistics}.
\newblock pages 72--79.

\bibitem[\protect\citename{Vaswani et al.}2017]{vaswani2017attention}
\newblock Ashish {Vaswani}, \xspace Noam {Shazeer}, \xspace Niki {Parmar}, \xspace Jakob {Uszkoreit}, Llion {Jones}, Aidan N. {Gomez}, Lukasz {Kaiser}, and Illia {Polosukhin}.
\newblock 2017.
\newblock Attention is All you Need.
\newblock {In \em Proceedings of the 31st International Conference on Neural Information Processing Systems}.
\newblock pages 5998--6008.

\bibitem[\protect\citename{Wang et al.}2012]{wang2012a}
\newblock Hao {Wang}, Dogan {Can}, Abe {Kazemzadeh}, François {Bar}, and Shrikanth {Narayanan}.
\newblock 2012.
\newblock A System for Real-time Twitter Sentiment Analysis of 2012 U.S. Presidential Election Cycle.
\newblock {In \em Proceedings of the ACL 2012 System Demonstrations}.
\newblock pages 115--120.

\bibitem[\protect\citename{Wibawa et al.}2013]{wibawa2013indonesian}
\newblock Aji~P. {Wibawa}, \xspace Andrew {Nafalski}, \xspace A.~Effendi {Kadarisman}, and Wayan~F. {Mahmudy}.
\newblock 2013.
\newblock {Indonesian-to-Javanese Machine Translation}.
\newblock {In \em International journal of innovation, management and technology}.

\bibitem[\protect\citename{Wibawa et al.}2018]{wibawa2018building}
\newblock Jaka Aris Eko {Wibawa}, Supheakmungkol {Sarin}, Chen Fang {Li}, Knot {Pipatsrisawat}, Keshan {Sodimana}, Oddur {Kjartansson}, Alexander {Gutkin}, Martin {Jansche}, and Linne {Ha}.
\newblock 2018.
\newblock {Building Open Javanese and Sundanese Corpora for Multilingual Text-to-Speech}.
\newblock {In \em Proceedings of the Eleventh International Conference on Language Resources and Evaluation (LREC-2018)}.

\bibitem[\protect\citename{Wu and Dredze}2019]{wu2019beto}
\newblock Shijie {Wu} and Mark {Dredze}.
\newblock 2019.
\newblock Beto, {B}entz, {B}ecas: The Surprising Cross-Lingual Effectiveness of {BERT}.
\newblock {In \em 2019 Conference on Empirical Methods in Natural Language Processing}.
\newblock pages 833--844.

\bibitem[\protect\citename{Xu et al.}2019]{xu2019bert}
\newblock Hu \xspace {Xu}, \xspace Bing \xspace {Liu}, \xspace Lei \xspace {Shu}, \xspace and Philip \xspace S. {Yu}.
\newblock 2019.
\newblock {BERT} Post-Training for Review Reading Comprehension and Aspect-based Sentiment Analysis.
\newblock {\em arXiv preprint arXiv:1904.02232}.

\bibitem[\protect\citename{Yasuda and Sumita}2008]{yasuda2008method}
\newblock Keiji {Yasuda} and Eiichiro {Sumita}.
\newblock 2008.
\newblock Method for Building Sentence-Aligned Corpus from Wikipedia.
\newblock {In \em 2008 AAAI Workshop on Wikipedia and Artificial Intelligence (WikiAI08)}.
\newblock pages 263--268.

\bibitem[\protect\citename{Yessenov and Misailovic}2009]{yessenov2009sentiment}
\newblock Kuat Yessenov and Sa{\v{s}}a Misailovic.
\newblock 2009.
\newblock Sentiment analysis of movie review comments.
\newblock {In \em Methodology}, 17: 1--7.

\bibitem[\protect\citename{Zhao and Vogel}2002]{zhao2002adaptive}
\newblock Bing {Zhao} and S. {Vogel}.
\newblock 2002.
\newblock Adaptive parallel sentences mining from web bilingual news collection.
\newblock {In \em Proceedings of 2002 IEEE International Conference on Data Mining, 2002}.
\newblock pages 745--748.



\end{thebibliography}
\end{document}